\documentclass[letterpaper]{article} 
\usepackage{aaai24}  
\usepackage{times}  
\usepackage{helvet}  
\usepackage{courier}  
\usepackage[hyphens]{url}  
\usepackage{graphicx} 
\usepackage{tabularx} 
\usepackage{array}
\usepackage{natbib}  
\usepackage{caption} 
\usepackage{algorithm}
\usepackage{algorithmic}
\usepackage{subfig}
\usepackage{multirow}
\usepackage{newfloat}
\usepackage{listings}
\DeclareCaptionStyle{ruled}{labelfont=normalfont,labelsep=colon,strut=off} 
\floatstyle{ruled}
\newfloat{listing}{tb}{lst}{}
\floatname{listing}{Listing}
\title{Enhancing Two-Player Performance Through Single-Player Knowledge Transfer: An Empirical Study on Atari 2600 Games}
\author{
    Kimiya Saadat,
    Richard Zhao
}
\affiliations{
    Department of Computer Science, University of Calgary\\
    Calgary, Alberta, Canada, T2N 1N4\\
    kimiya.saadat@ucalgary.ca, richard.zhao1@ucalgary.ca
}

\begin{document}

\maketitle

\begin{abstract}
Playing two-player games using reinforcement learning and self-play can be challenging due to the complexity of two-player environments and the possible instability in the training process. We propose that a reinforcement learning algorithm can train more efficiently and achieve improved performance in a two-player game if it leverages the knowledge from the single-player version of the same game. This study examines the proposed idea in ten different Atari 2600 environments using the Atari 2600 RAM as the input state. We discuss the advantages of using transfer learning from a single-player training process over training in a two-player setting from scratch, and demonstrate our results in a few measures such as training time and average total reward. We also discuss a method of calculating RAM complexity and its relationship to performance.
\end{abstract}

\section{Introduction}
In recent years, the advancement of artificial intelligence (AI) research in deep neural networks and reinforcement learning (RL) has shown prominent results in playing different video games, especially games in Atari 2600 (referred to as Atari in the rest of this paper). The first deep Q network (DQN) was evaluated on seven Atari environments \cite{mnih2013playing}. RL has been used in single-player and multiplayer games. Despite the achievements, deep RL algorithms have been shown to be extensively sample-inefficient, especially when using high-dimensional input data \cite{yarats2021improving}. DQN on Atari using images takes millions of steps to learn to play a game \cite{mnih2013playing}.

Alongside the general complexity of mapping observations to a singular reward, more problems can arise when using self-play. When training a multi-player game, a technique called self-play can be employed, where the agent plays against or in collaboration with a generation of itself. An example of this can be seen in AlphaGo Zero, which is an evolution of AlphaGo, an AI system that was able to defeat a human champion in the board game Go \cite{silver2017mastering}. Using self-play in multi-player environments can be challenging due to the environment being non-stationary. In other words, in a self-play setting, the opponent can be assumed to be a part of the environment. As the policy of the opponent changes, the environment becomes non-stationary and an action's effects may depend on the actions of other agents \cite{chakraborty2014multiagent}. This could hurt the possibility of convergence of self-play in a multi-player environment \cite{hammar2020finding}. 

Transfer learning or knowledge transfer is a technique in which the knowledge gained from a different domain is transferred to the target domain to help the process of learning \cite{zhuang2020comprehensive}. Transfer learning has been used extensively in RL. Examples include the use of transfer learning for gaining the ability to generalize between multiple environments \cite{parisotto2015actor} and solving multi-task problems \cite{rusu2016progressive}. Particularly, a specific use of transfer learning is in curriculum learning \cite{da2019survey}, where the complex final task is learned faster by breaking the task into simpler tasks in the same domain. In our case, we have a similar goal, starting from a simpler single-player environment and moving to a more complex one where two agents have to play together. However, the specific use of single-player version of a game to train the two-player version of the same game has not been explored by other studies and it is not clear whether that would introduce an advantage in the process of learning.  

Our research questions are: 1. Can knowledge learned from a simpler single-agent Atari environment be transferred to the corresponding two-player environment of the same game, enabling an agent to outperform one trained exclusively in the two-player environment? 2. Is there a relationship between the utilization of the Atari RAM and the performance of the transferred agent? Could we use the complexity of RAM utilization to predict transferred agent performance? In this work:
\begin{enumerate}
\item We show that leveraging knowledge transfer from a single-player Atari environment can have advantages over training from scratch in a two-player environment.
\item We propose a method of quantifying the complexity of the Atari RAM state for a particular game, along with a visualization of such complexity.
\item We discuss how RAM complexity is correlated with the performance of the transferred agent and can be used as a weak predictor.
\end{enumerate}

\section{Related Works}
\subsection{Transfer Learning and RL}

Transfer learning refers to using knowledge gained from one or multiple source domains to improve a learner in a target domain. Transfer learning strategies are applied to both traditional machine learning techniques and deep neural networks. Using transfer learning could result in more efficient training and better performance, since the source domain(s) has already useful features that could be leveraged for the new task \cite{zhuang2020comprehensive}.

Reinforcement learning is a subfield of machine learning in which an agent aims to maximize the cumulative reward while interacting with its environment \cite{sutton2018reinforcement}. Utility values for states or Q values for state-action pairs are updated through trial and error. RL can be used to train agents that play games such as chess, poker, or Go at levels that surpasses the capabilities of the human players or is comparable to the best of them. \cite{silver2016mastering}. A DQN uses a deep neural network to approximate the Q values and is able to handle complex learning environments such as those found in video games by using the entire screen of pixels as inputs \cite{justesen2019deep}. The use of DQN and its successors have allowed researchers to tackle a vast set of video games, from first-person shooters to sports games \cite{justesen2019deep}. Deep RL approaches, however, are not without drawbacks: among them, interpretability \cite{eberhardinger2023learning, sieusahai2021explaining} and long training time.

Combining transfer learning and RL allows for quicker learning and learning more complex problems \cite{braylan2016object}. This concept has been applied to a relational regression algorithm used to learn a generalized Q-function \cite{ramon2007transfer}. More recently, Gamrian and Goldberg (\citeyear{gamrian2019transfer}) show that transfer knowledge between slightly modified RL environments can be improved using Generative Adversarial Networks (GANs) with imitation learning. Anwar and Raychowdhury (\citeyear{anwar2020autonomous}) demonstrate the effectiveness of transfer learning in deep RL for autonomous navigation in the Unreal game engine. Pleines et al. (\citeyear{pleines2022verge}) propose sim-to-sim transfer, transferring knowledge learned in a limited Unity engine implementation of Rocket League to the full game, showing that knowledge quickly learned in a simplified environment can be useful in a more complex environment. Balla and Perez-Liebana (\citeyear{balla2022task}) present research on the effectiveness of pre-training Successor Features in transferring their knowledge to new target tasks without further training. The survey by Muller-Brockhausen et al. (\citeyear{muller2021procedural}) discusses the difficulty of transfer learning in generalizing to different problem variations and the need for a united benchmark to test transfer learning in RL. This is an ongoing challenge requiring further studies. 

\subsection{Multi-Agent Environments}
A multi-agent environment refers to a setting where multiple autonomous agents learn to interact with each other and the environment simultaneously. This framework is often used to model complex, dynamic systems with multiple decision-makers such as multiplayer games. In a multi-agent environment, each agent has its own goals and learning process. They may cooperate, compete, or do both, depending on the situation \cite{tampuu2017multiagent}. Their actions affect not only their own future state but also those of the other agents, creating a complex, intertwined learning problem.

There are many different approaches in multi-agent environments. Boeda (\citeyear{boeda2021multi}) proposes planning using Goal Oriented Action Planner to facilitate cooperation among multiple agents. Lei and Zhang (\citeyear{lei2017service}) study cooperative game theory with a service composition method.

One common method for training agents in multi-agent environments is RL. Multi-Agent RL (MARL) extends RL to multi-agent settings, where multiple agents learn to optimize their behaviors while interacting with each other. Chakraborty and Stone (\citeyear{chakraborty2014multiagent}) examine convergence to Nash equilibrium and targeted-optimality in learning. Booth and Booth (\citeyear{booth2019marathon}) present Marathon Environments, a suite of continuous control benchmarks for multi-agent RL in the Unity game engine. Ferreira et al. (\citeyear{ferreira2022optimizingmarl}) propose a framework for developing multi-agent cooperative game environments to facilitate the process and improve agent performance during RL.  MARL has also been examined in StarCraft \cite{khan2022transformer}, where a transformer-based joint action-value mixing network is shown to be superior to other benchmarks.

Won, Gopinath, and Hodgins (\citeyear{won2021control}) present a learning framework that generates control policies for physically simulated athletes in two-player games, boxing and fencing. The framework uses a two-step approach for learning basic skills and learning bout-level strategies. The authors develop a policy model based on an encoder-decoder structure that incorporates an autoregressive latent variable and a mixture-of-experts decoder. The pre-trained model with basic skills is transferred over and trained on the two-player environment.

In our work, we explore the use of knowledge transfer from a single-agent environment to its corresponding multi-agent environment, sidestepping the complications from training directly in a multi-agent environment. Training in a single-agent environment has several potential advantages over a multi-agent environment, mainly due to its stablility and efficiency. Single-agent environments are often more stable and predictable because there is no other learning entity to introduce non-stationarity. In multi-agent settings, each agent's policy is changing as it learns, which can lead to a constantly shifting environment that is challenging to learn from.  Training can be more computationally efficient in single-agent settings. Each additional agent in a multi-agent setting increases the complexity of the state and action spaces, which can slow down learning.

We deploy the concept of self-play in our empirical study. Self-play in RL is defined to be a method of deploying an algorithm against copies of itself to learn and test in various stochastic learning environments. Self-play has been well-tested in many applications. AlphaGo Zero and AlphaZero \cite{silver2018general} famously use self-play to achieve superhuman performance in chess, shogi, and Go. Self-play has also seen success in fighting game AI \cite{takano2019self}, ``Big 2", a four-player
game of imperfect information \cite{charlesworth2019application}, among others. Liu et al. (\citeyear{liu2021self}) introduce a critic into
the policy gradient method to form a self-play actor-critic method for training agents and demonstrate its success in a few simple environments.

\subsection{The Atari Environments}

In our empirical study, we have chosen ten different Atari environments provided by Gymnasium (single player) \cite{towers2024gymnasium} and PettingZoo (two players) \cite{terry2021pettingzoo}. Both of these libraries use the Arcade Learning Environment (ALE) which is a platform for building intelligent agents across different Atari games \cite{shao2019survey}. There have been numerous attempts at training AI agents to play Atari games using RL. One of the most notable achievements is when DQN reached human-level performance across 49 Atari games \cite{DQNpaper}. After this breakthrough, other algorithms and improvements of DQN such as Rainbow DQN and Ape-X DQN have shown promising results and in some cases surpassing DQN results \cite{shao2019survey}. More recently, a novel approach of creating an auxiliary reward using instructions extracted from the manuals of the Atari games have shown  improvement in performance and training speed \cite{wu2024read}. For the purpose of this study, we focus on the regular DQN algorithm that is common with Atari games. We use DQN with a couple of improvements such as prioritized experience replay and double structure to aid the training process.

\section{Methodology}
We begin by describing the environments used for training. Next, details of the RL algorithm are provided, alongside the transfer method. The final aspect we discuss is the configuration of the experiments we conducted to investigate the advantages of knowledge transfer. 

\subsection{Details of the Environments}

For developing our Atari transfer benchmark, we focus on the games that are present in both Gymnasium and PettingZoo libraries, with single-player and two-player versions. Gymnasium provides single agent Atari environments while PettingZoo offers a library to facilitate multi-agent Atari training. Atari environments are simulated via the Arcade Learning Environment. We use a total of nine Atari games in ten Atari environments. The names of the environments sorted alphabetically are: Boxing, Double Dunk, Entombed competitive, Entombed cooperative, Flag Capture, Mario Bros, Pong, Space Invaders, Surround, and Tennis. Entombed is a game with two different environments, cooperative and competitive, based on the strategy of the player. Throughout the rest of this paper we use ``Atari games" as a shorthand for Atari game environments. For training, the Atari RAM is used instead of image pixels to optimize computational power and time. RAM observations are a total of 128 bytes.  

In single-player environments, all games except Mario Bros, Entombed, Flag capture and Surround use version 4 (v4) environments with default frame skip of one and no action repeating. The four mentioned games use version 5 as it yields better results. In version 5 there is a default frame skip of 4 and repeat action probability of 0.25. 

Single-player training is conducted using the Tianshou platform \cite{tianshou}. In addition to the environment's settings, the default processing of the DQN Atari benchmark in Tianshou is used. These include sampling initial states by taking random number of no-ops on reset, returning every 4th frame (frame skipping) with maxing over 2 most recent raw observations, making the end of life the same as end of episode (bootstrapping values) without resetting to help with the value estimation, normalizing the observation between 0 and 1, and clipping the reward to be between -1 and 1.

For two-player environments, the same reward clipping and observation normalization is used to ensure that the network can comprehend values after transfer. Based on the work of Lee et al. (\citeyear{lee_investigation_2022}), we also add the following preprocessing techniques: 
\begin{enumerate}
\item Limiting the max number of steps in each episode to 200.
\item Frame skip of 4.
\item Sticky actions with a probability of 0.25.
\end{enumerate}
Pre-processing in two-player Atari games is done using the SuperSuit library \cite{SuperSuit} and training is done using the Machin library \cite{machin}.
In two-player training, no-op resets are performed on the first 130 frames for Space Invaders, and the first 60 frames for Pong \cite{lee_investigation_2022}.
Two-player environments reset upon termination of first-player. In Mario Bros, the game can continue even if player two is dead, so we ensure that the game resets when player two is terminated as well.

\begin{figure*}[htb]
\centering
\captionsetup{labelformat=empty}
\captionsetup[subfigure]{}

    \subfloat[Double Dunk]{
        \includegraphics[width=0.23\textwidth]{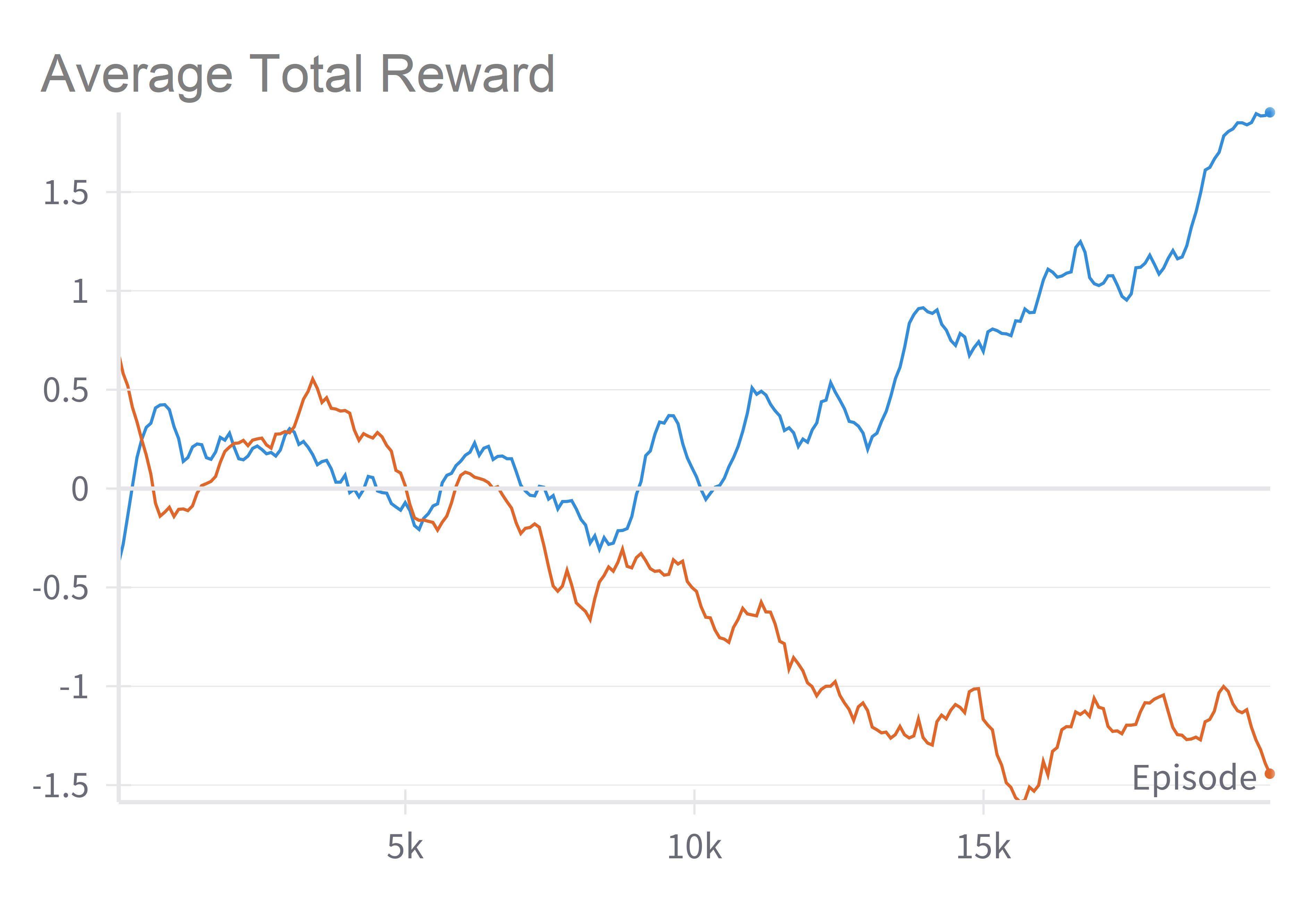}
        \label{fig:figure1}
    }
    \hfill 
    \subfloat[Mario Bros]{
        \includegraphics[width=0.23\textwidth]{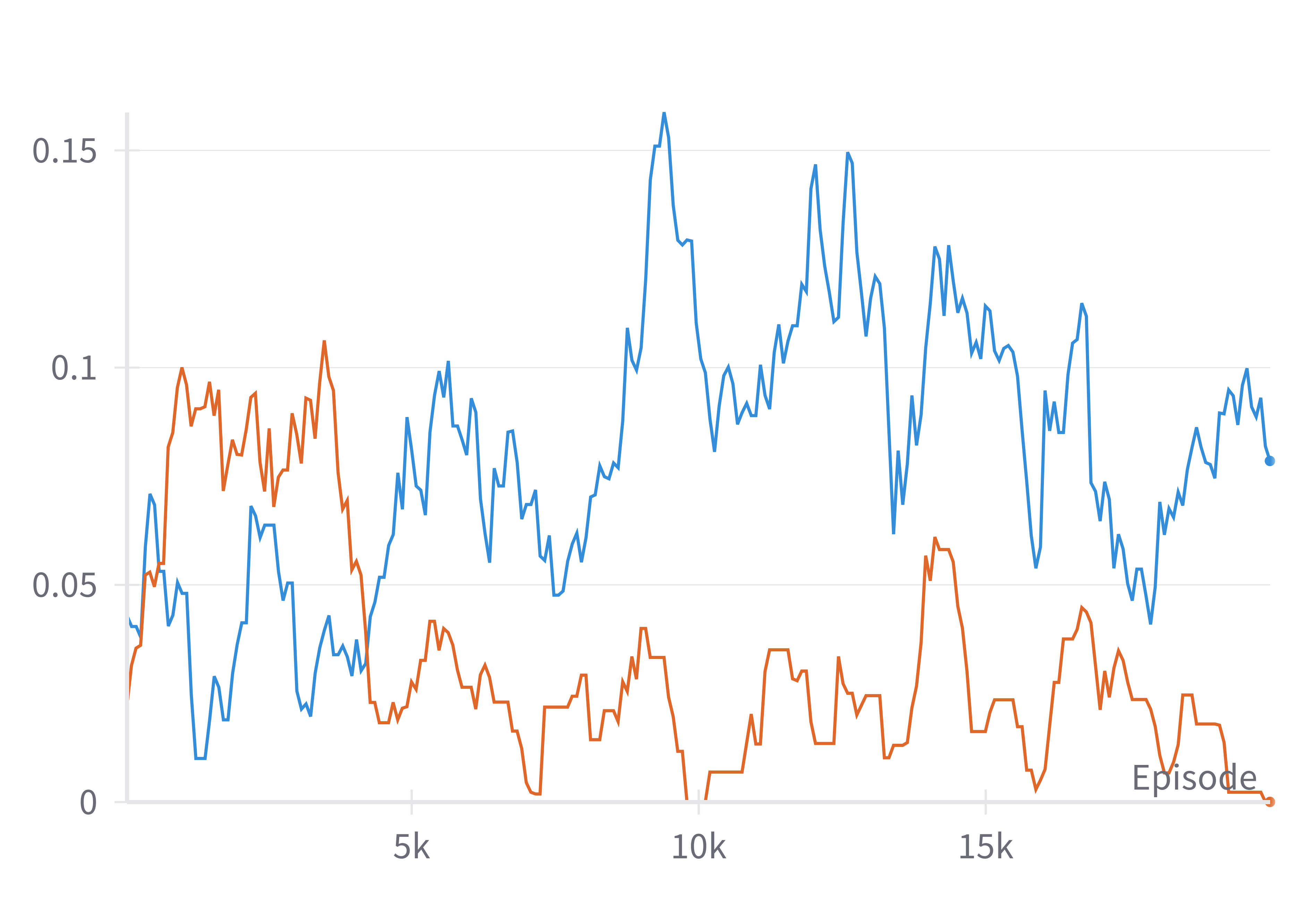}
        \label{fig:figure7}
    }
    \hfill 
    \subfloat[Space Invaders]{
        \includegraphics[width=0.23\textwidth]{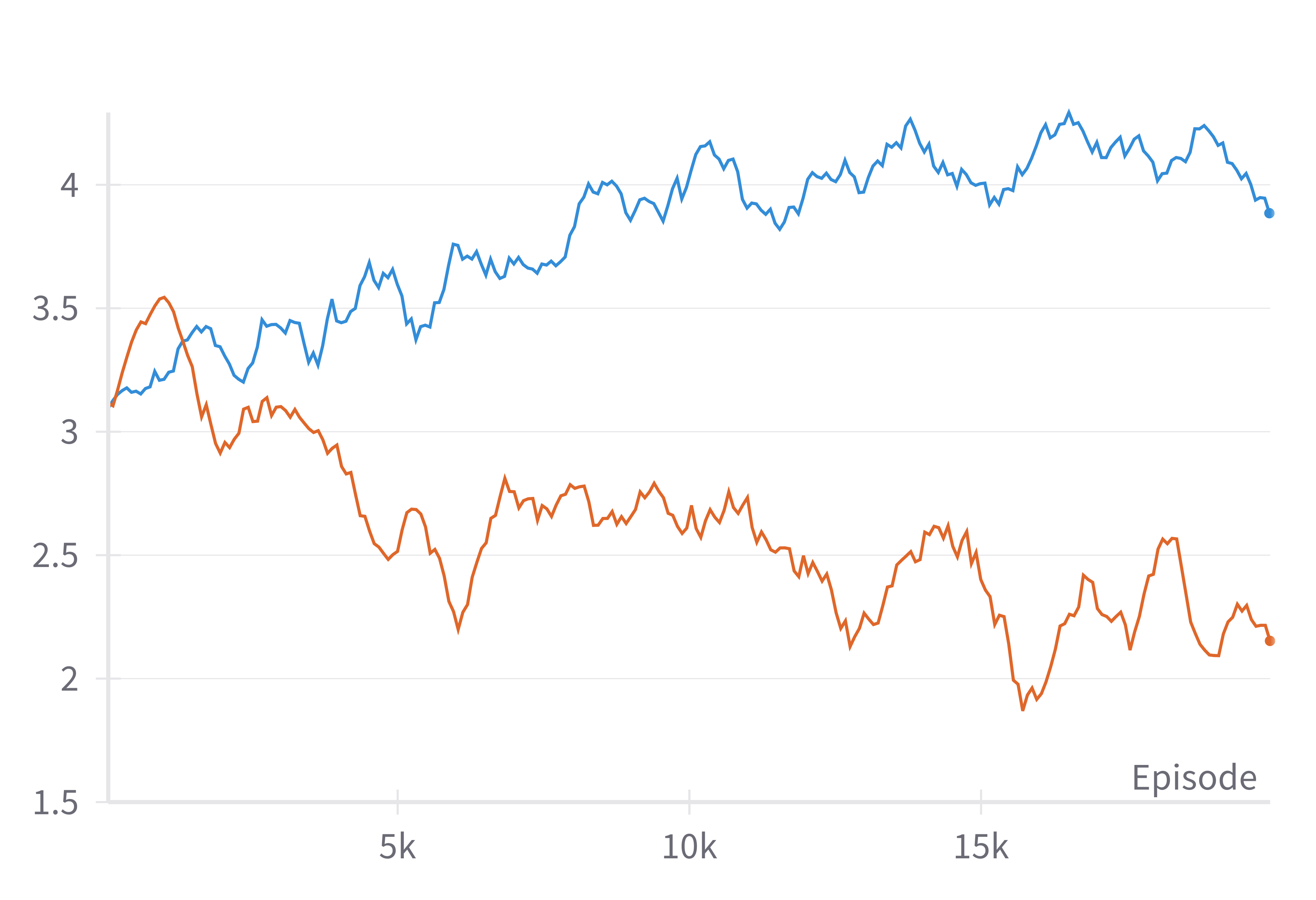}
        \label{fig:figure2}
    }
    \hfill 
    \subfloat[Tennis]{
        \includegraphics[width=0.23\textwidth]{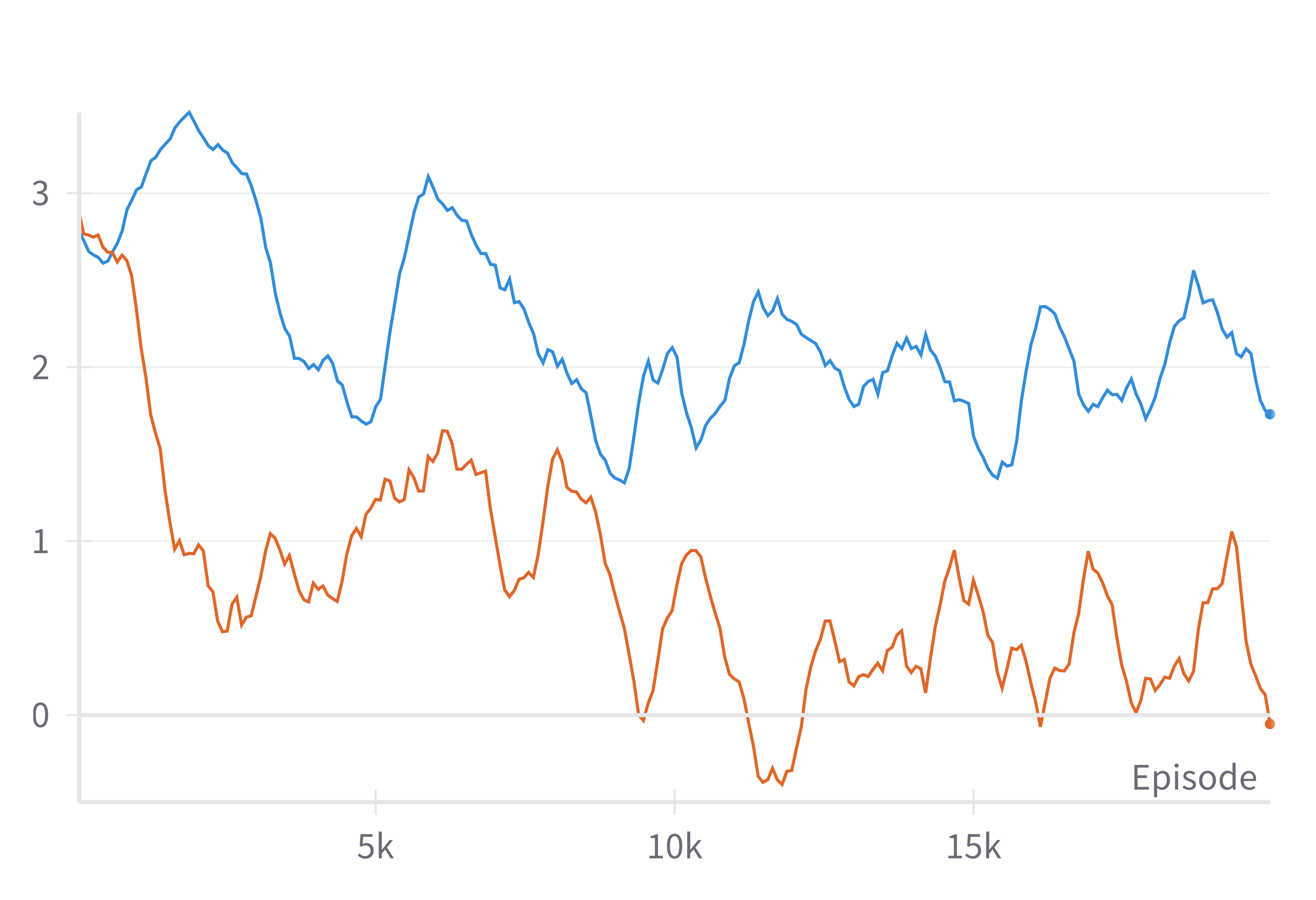}
        \label{fig:figure3}
    }
    
    \subfloat[Boxing]{
        \includegraphics[width=0.23\textwidth]{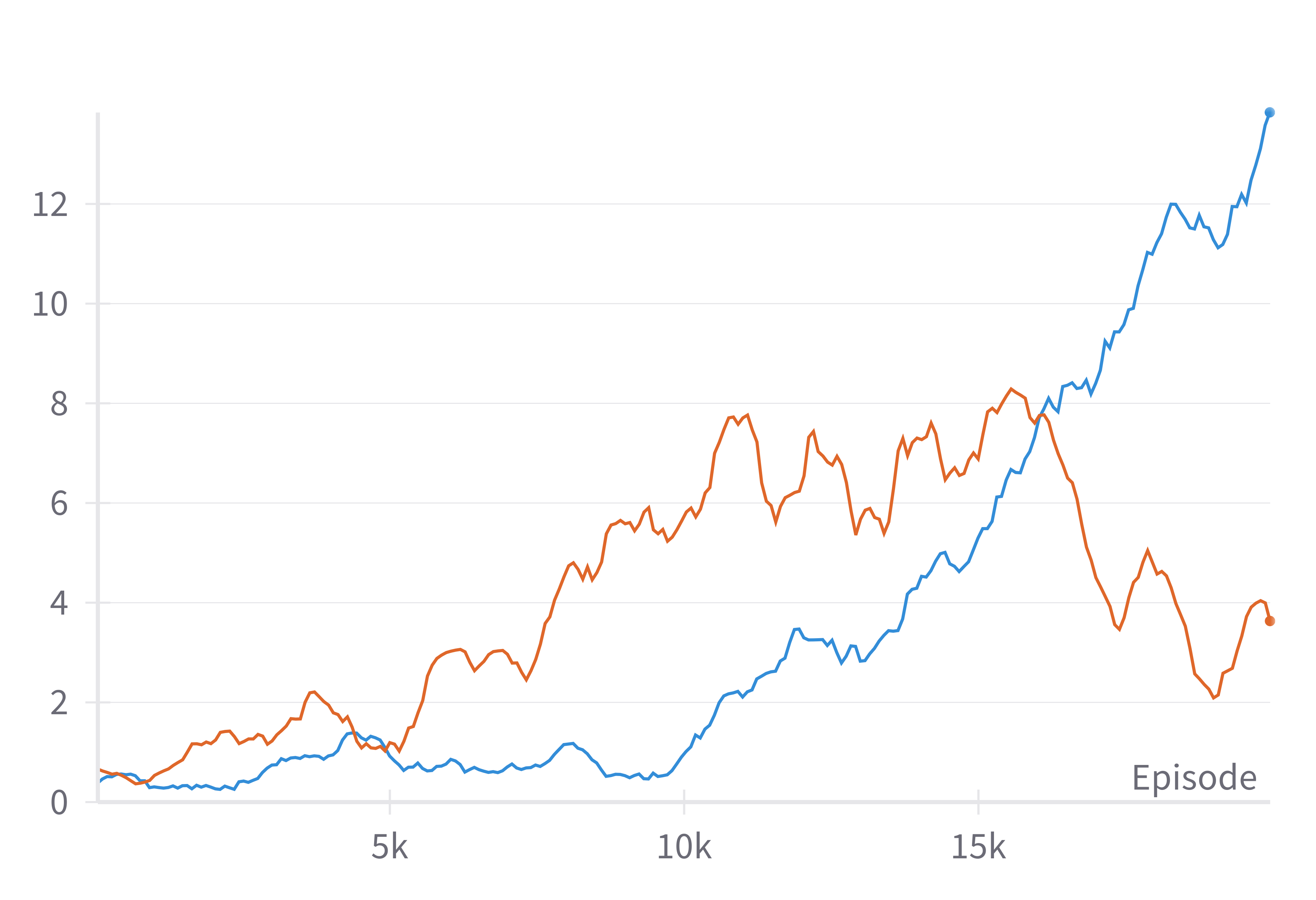}
        \label{fig:figure4} 
    }
    \hfill 
    \subfloat[Pong]{
        \includegraphics[width=0.23\textwidth]{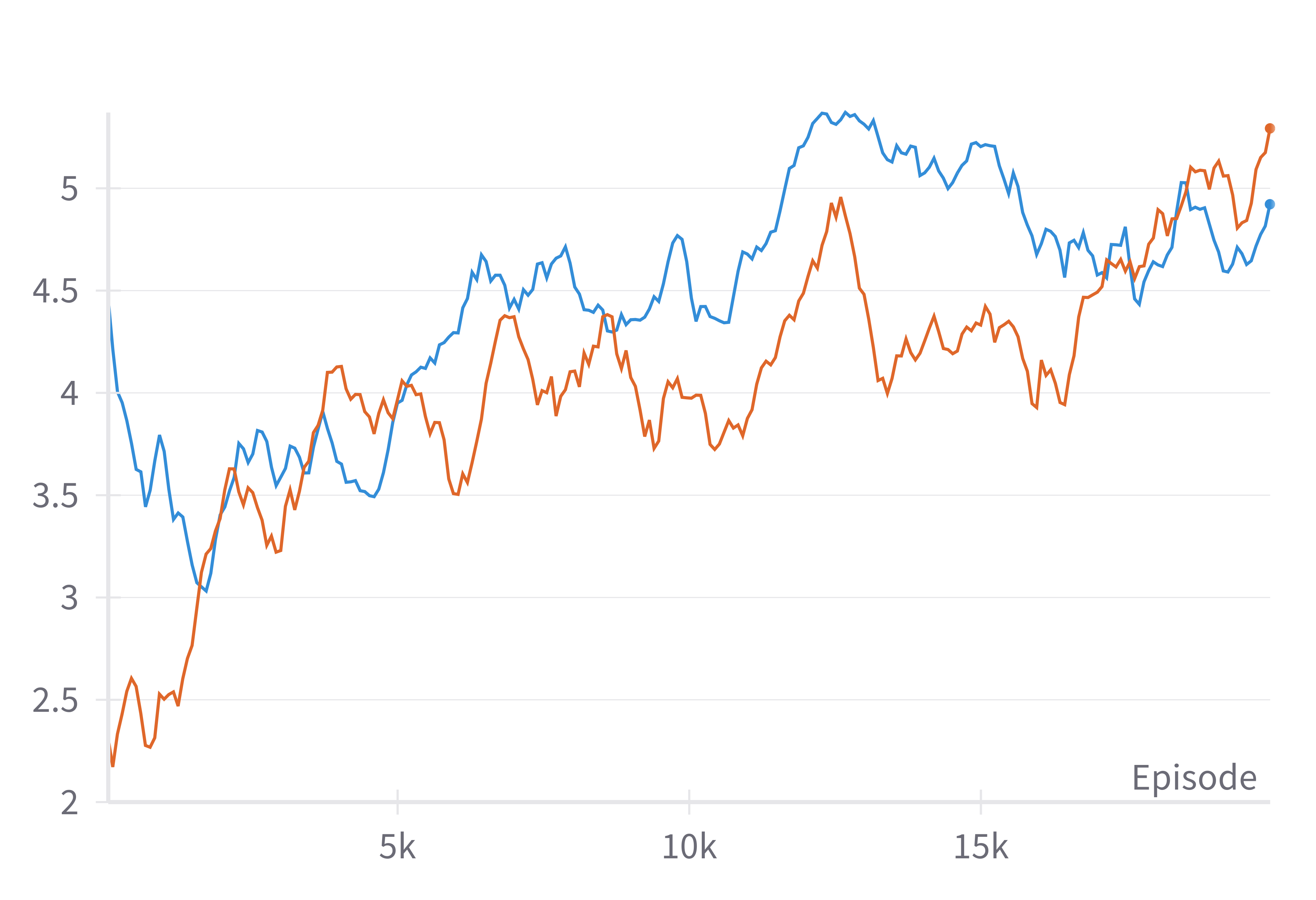}
        \label{fig:figure5}
    }
    \hfill 
    \subfloat[Flag Capture]{
        \includegraphics[width=0.23\textwidth]{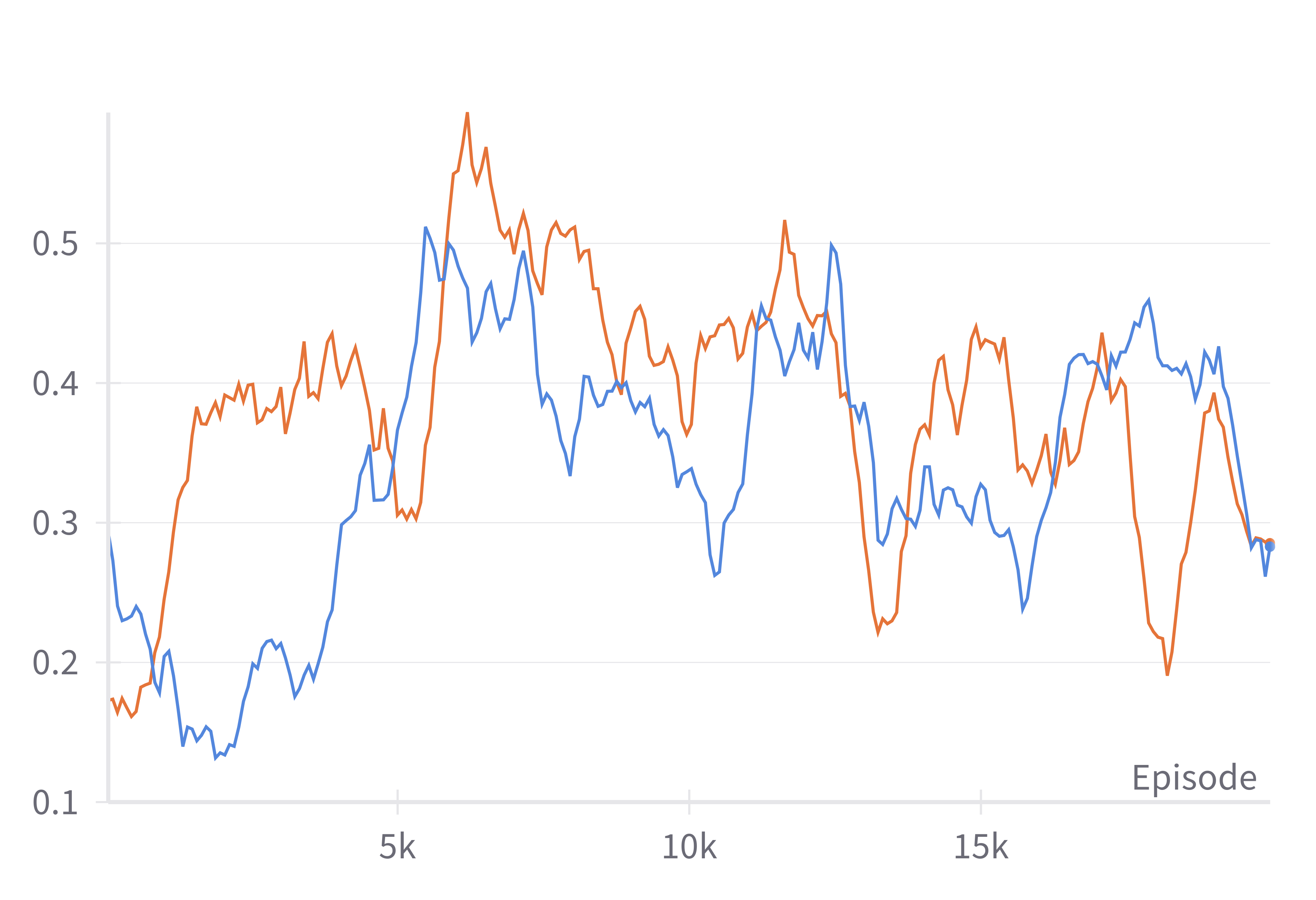}
        \label{fig:figure6}
    }
    \hfill 
    \subfloat[Entombed Competitive]{
        \includegraphics[width=0.23\textwidth]{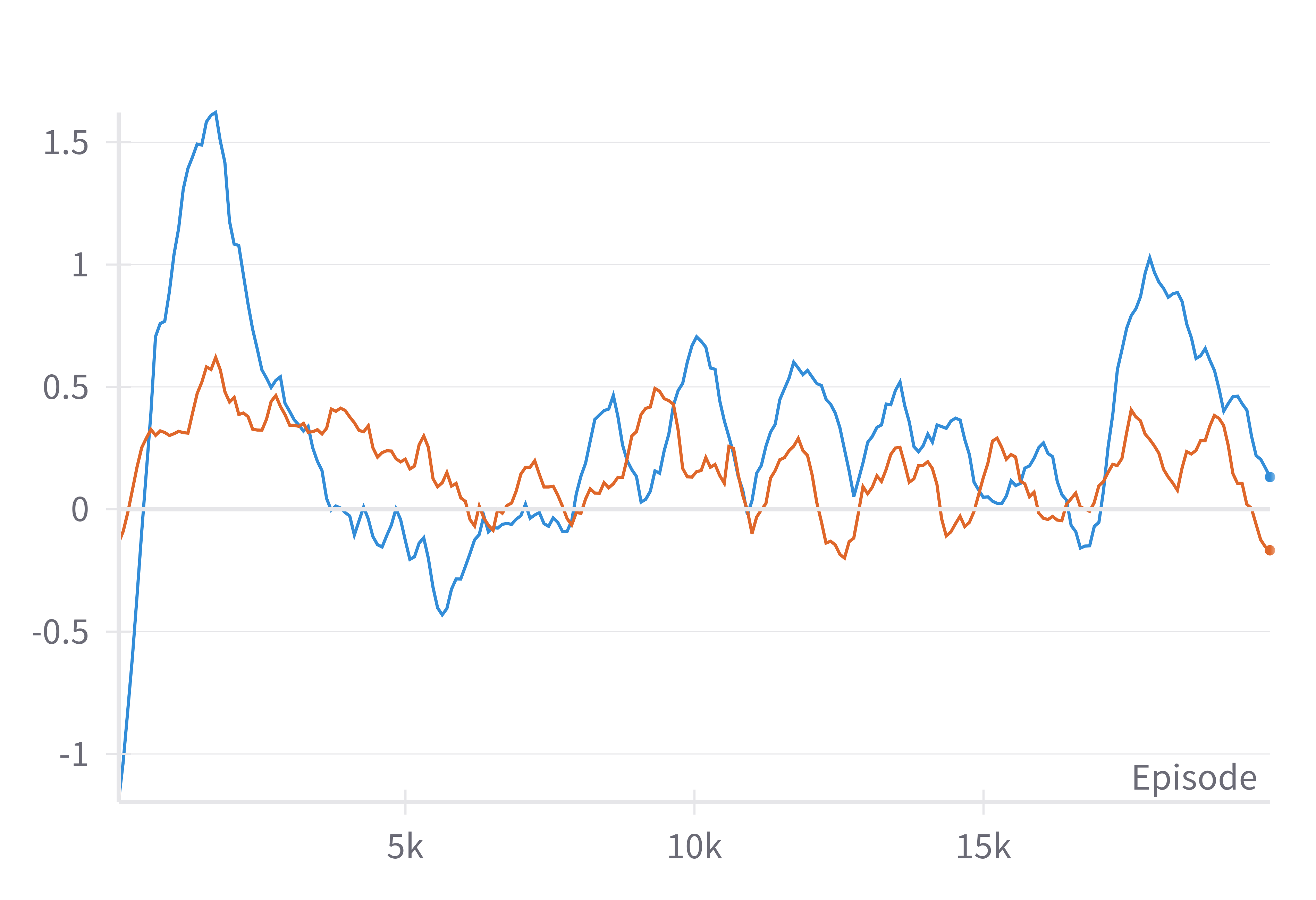}
        \label{fig:figure8}
    }

    \subfloat[Entombed Cooperative]{
        \includegraphics[width=0.23\textwidth]{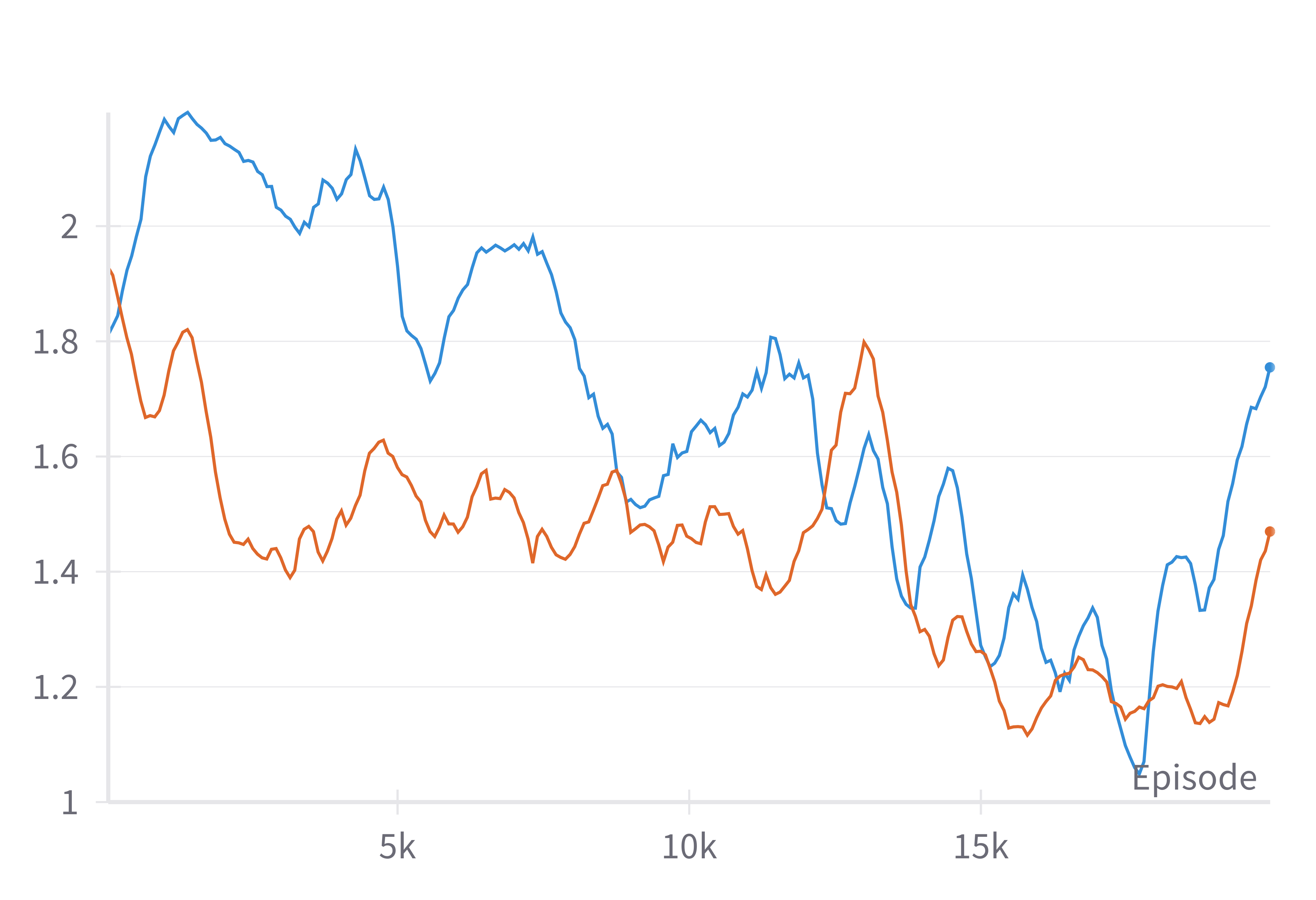}
        \label{fig:figure9}
    }
    \hfill 
    \subfloat[Surround]{
        \includegraphics[width=0.23\textwidth]{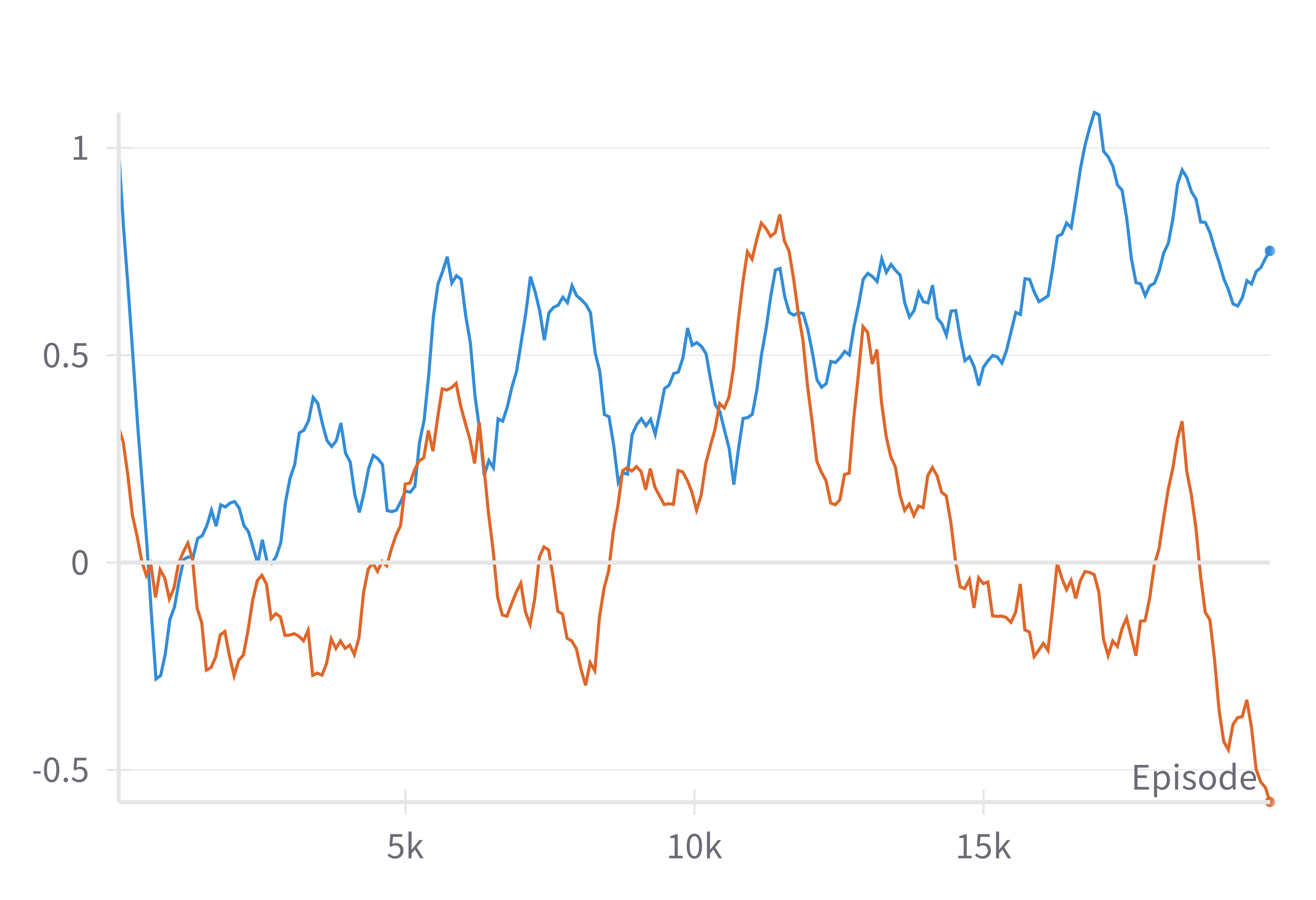}
        \label{fig:figure10}
    }
    
    \caption{Figure 1: The Average total reward of player 1 in the ten Atari games for the transferred agent (in blue) and the agent training from scratch (in orange), in the two-player versions of the games. The x-axis is the number of episodes and the y-axis is the average total reward per episode.}
    \label{fig:figure1comb}
\end{figure*}

\subsection{The Algorithm}

A double deep Q-network is the main RL algorithm to train on the games for both single-player and two-player environments. Prioritized experience replay is used alongside this to ensure that experiences with a higher importance get prioritized during training. A fully connected deep neural network with the same structure for both single-player and two-player games is used to map observations of 128 bytes to the number of actions that varies based on the environment but is consistent for both single-player and two-player versions of the same game. The details of this neural network and hyper-parameters used for training can be found in Table \ref{table:parameters}.
\begin{table*}[htb]
  \begin{center}
    \begin{tabular}{c|c|c}
    \hline 
      Hyperparameter & Value for single-player & Value for two-player \\
      \hline
      fully-connected layer dimensions & 512×256 & 512×256 \\
      
      number of total steps & 10,000,000 & Maximum of 4,000,000 \\
      
      optimizer & Adam & Adam \\
      
      discount factor & 0.99 & 0.99 \\
      
      learning rate & 0.0001 & 0.001 \\
      
      batch size & 32 & 256 \\
      
      reply buffer size & 100,000 & 500,000 \\
      
      epsilon decay rate & N/A (linear decay from 1 to 0.05) & 0.9999985 \\ 
      
      number of parallel environments & 10 & 1 \\
      
      player2 fixed exploration rate & N/A & 0.05 \\
      
      seeds & 99 & [24, 42, 56, 99, 3000] \\
      
      self-play step & N/A & 50000 \\
      
       \hline 
    \end{tabular}
  \end{center}
      \caption{Hyperparameters used in training for both single-player and two-player games.}
    \label{table:parameters}
\end{table*}
For playing the opponent in two-player versions of games, self-play is used. During training, only the network of player 1 (also called ``the player") is trained while the network of player 2 (also called ``the opponent") is only updated at certain time-steps by copying the weights of player 1's network. This ensures that the opponent's network is always updated with the previous generations of the player's network. The network of player 1 is solely trained using player 1's experience. Since we aim to not train the opponent's network separately, we need to ensure that the network understands the difference between players 1 and 2 when playing. For this purpose, we use our method of agent indication in two-player Atari that is discussed in the next section.

\subsection{Agent Indication}

In Atari, all agents have the same observation from the environment, which could be a problem when there is a need to indicate which player is which. A common method of agent indication is by adding a separate channel to the observation to indicate the agent's turn \cite{gupta2017cooperative}. However, adding a new channel to a network could make the training more difficult as it could make it harder for transfer learning to discover parts of the observation that are completely new. To avoid this problem, we annotate the parts in the Atari two-player RAM that are directly related to player 1 or 2. The annotation has been done using both PettingZoo and PCAE Atari emulator. Annotations of four environments are inspired by the annotations from the Atari Annotated RAM Interface \cite{anand2019unsupervised}. Our focus is mainly on elements that could affect a player's gameplay such as player avatars' locations, scores, and the number of remaining lives. After annotating the desired parts, we implement a PettingZoo custom wrapper that would reconstruct the observation of player 2 by swapping the related parts of player 2 with player 1. As an example, the bytes related to locations of player 1 and 2 are swapped. Meaning that the specific byte that contains the location of player 1 now contains the location of player 2 and the network is able to decide based on this location. Essentially, the observation of player 2 is reconstructed as if player 1 were the one playing instead.
The code used for training two-player environments including this custom wrapper is publicly available in our GitHub repository at \url{https://github.com/Justkim/Two-player-atari-RL} . 

\begin{table*}[htb]
\centering 
\begin{tabular}{l|c |c |c |c |c c} 
\hline 
\hspace{2cm} Episode & 1 & 5000 & 10000 & 15000 & 20000 \\ [0.5ex] 
Game &  &  &  &  &  \\ [0.5ex] %
\hline 
\multirow{ 2}{*}{Double Dunk} & -1.0 ± 1.78 & 0.4 ± 1.85 & -0.4 ± 1.85 & 1.2 ± 1.46 &  2.0 ± 2.09\\
& 1.2 ± 1.16 & 0.4 ± 0.8 & -1.0 ± 0.89 & -1.4 ± 1.49 & -0.4 ± 0.8 \\
 \hline
  \multirow{ 2}{*}{Mario Bros} & 0.0 ± 0.0 & 0.0 ± 0.0 & 0.2 ± 0.4 & 0.0 ± 0.0 & 0.4 ± 0.48 \\ 
& 0.0 ± 0.0 & 0.0 ± 0.0 & 0.0 ± 0.0 & 0.0 ± 0.0 & 0.0 ± 0.0\\
 \hline
 \multirow{ 2}{*}{Space Invaders} & 2.8 ± 1.6 & 3.4 ± 1.62 & 3.6 ± 1.49 & 5.6 ± 2.15 & 4.0 ± 1.0 \\ 
& 3.4 ± 1.01 & 2.0 ± 1.09 & 2.8 ± 1.59 & 3.2 ± 2.4 & 1.5 ± 0.5\\
 \hline
 \multirow{ 2}{*}{Tennis} & 2.2 ± 1.83 & 1.6 ± 2.33 & 1.2 ± 1.93 & 0.4 ± 1.35 & 2.4 ± 1.35 \\ 
& 4.0 + 0.0 & -0.4 ± 2.87 & 1.6 ± 2.93 & -0.2 ± 3.48 & -0.8 ± 3.24 \\[1ex] 
 \hline
\multirow{2}{*}{Boxing} & 2.0 ± 2.28 & 0.8 ± 1.72 & 1.4 ± 2.8 & 4.8 ± 8.51 & 6.8 ± 4.4 \\ 
& 1.6 ± 2.41 &-1.0 ± 2.6 & 6.6 ± 7.73 & 6.4 ± 10.17 & 3.8 ± 3.81\\
 \hline
\multirow{ 2}{*}{Pong} & 3.6 ± 3.32 & 4.6 ± 2.33 & 6.0 ± 0.0 & 6.0 ± 0.0 & 5.0 ± 1.26\\ 
& 2.8 ± 3.65 & 3.8 ± 1.32 & 3.4 ± 3.55 & 2.6 ± 3.26 & 5.2 ± 1.6\\
 \hline
 \multirow{ 2}{*}{Flag Capture} & 0.0 ± 0.0 & 0.4 ± 0.48 & 0.4 ± 0.48 & 0.2 ± 0.4 & 1.0 ± 0.81 \\ 
& 0.0 ± 0.0 & 0.6 ± 0.48 & 0.6 ± 0.48 & 0.4 ± 0.48 & 0.4 ± 0.48\\
 \hline
\multirow{ 2}{*}{Entombed Competitive} & -1.2 ± 0.74 & 0.0 ± 1.41 & 0.6 ± 2.05 & 0.2 ± 0.97 & 0.2 ± 0.74 \\
& -0.2 ± 2.03 & -0.2 ± 1.72 & -0.4 ± 0.48 & -0.2 ± 1.16 & 0.0 ± 0.63\\
 \hline
\multirow{ 2}{*}{Entombed Cooperative} & 1.4 ± 0.48 & 2.0 ± 0.0 & 1.2 ± 0.97 & 1.0 ± 0.63 & 1.8 ± 0.4 \\
& 2.0 ± 0.0 & 1.6 ± 1.2 & 1.6 ± 0.8 & 1.0 ± 0.0 & 1.0 ± 0.63\\
 \hline
\multirow{ 2}{*}{Surround} & 0.4 ± 1.35 & -0.6 ± 0.79 &  0.2 ± 1.32 & 0.0 ± 0.63 & 1.4 ± 0.48 \\ 
& -0.4 ± 1.35 & -0.2 ± 1.32 & 1.0 ± 1.54 & -0.2 ± 0.97 & -0.4 ± 1.49 \\
 \hline

 \hline 
\end{tabular}
\label{table:means} 
\caption{Mean and standard deviation of total reward per episode, at the specified episodes. For each game, the first row represents transferred agent and the second row represents the agent training from scratch.} 
\end{table*}

\subsection{Transfer Learning}

For the transfer of knowledge from a single-player game to the two-player version of the same environment, the agent is first trained in single-player setting. After this training, all the weights are transferred to a new deep neural network to be trained in a two-player setting. To make the overall training more efficient and decrease the total running time while leveraging knowledge from the transferred network, the first two layers are frozen and do not get updated during two-player training. We have conducted preliminary experiments using different numbers of layer freezing, and found that freezing the first two layers produced the most desirable results.

\subsection{Experiment Setup}

As part of our research, we conduct an empirical study in the ten Atari games. For each game, two sets of training are conducted. For the transferred agents, first the network is pre-trained on the single-player game for 10 million steps. Then, the weights of this network is transferred to the two-player version of the same game for further running of 20,000 episodes. An episode refers to a complete sequence of steps that starts from an initial state and ends when the environment reaches a terminal state. Five different sessions with different random seeds are used in the two-player environments. For the training-from-scratch agents, five sessions with different random seeds in each game are done from scratch in the two-player version without any transfer for 20,000 episodes. The results of these experiments in the two-player versions are compared (transferred vs. from scratch) using the rewards gathered by the player in each episode. Across the different games and experiments, the same hyper-parameters are used to ensure consistency and avoid bias. 

Table \ref{table:parameters} reports the set of hyper-parameters for training single-player and two-player environments. The experiments (two-player training) were conducted using a cluster with 7 cores of 2x Intel(R) Xeon(R) Gold 5320 CPU @ 2.20GHz (Ice Lake) and a maximum of 10G RAM.

\section{Results and Discussions}

\begin{figure*}[htb]
\centering
\captionsetup{labelformat=empty}
\captionsetup[subfigure]{}

    \subfloat[Double Dunk]{
        \includegraphics[width=0.17\textwidth]{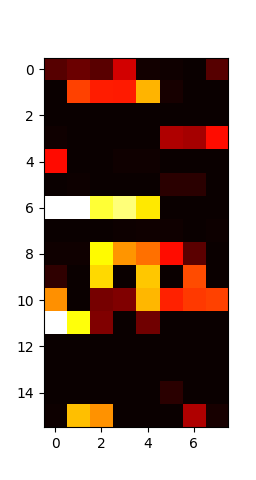}
        \label{fig:heatmap1}
    }
    \hfill 
        \subfloat[Mario Bros]{
        \includegraphics[width=0.17\textwidth]{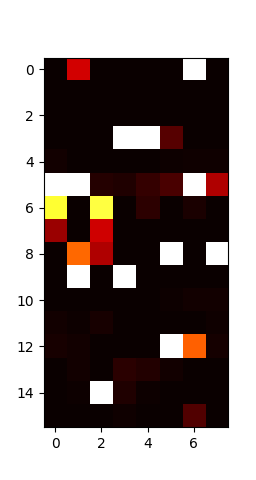}
        \label{fig:heatmap7}
    }
    \hfill 
    \subfloat[Space Invaders]{
        \includegraphics[width=0.17\textwidth]{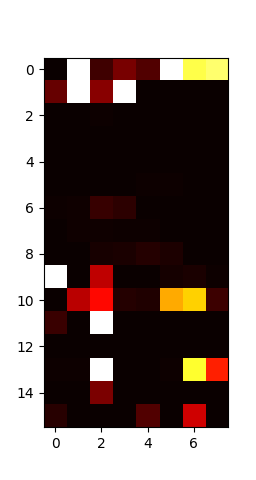}
        \label{fig:heatmap2}
    }
    \hfill 
    \subfloat[Tennis]{
        \includegraphics[width=0.17\textwidth]{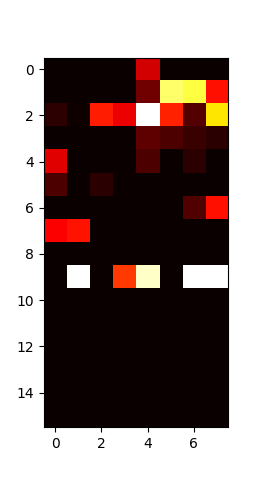}
        \label{fig:heatmap3}
    }
    \hfill 
    \subfloat[Boxing]{
        \includegraphics[width=0.17\textwidth]{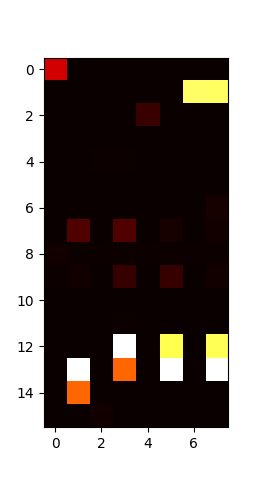}
        \label{fig:heatmap4} 
    }

    \subfloat[Pong]{
        \includegraphics[width=0.17\textwidth]{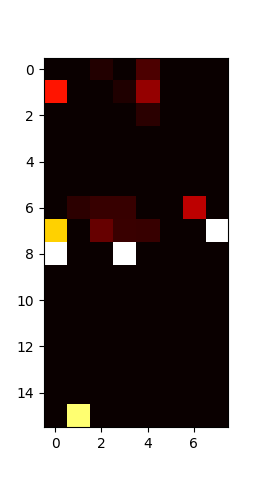}
        \label{fig:heatmap5}
    }
    \hfill 
    \subfloat[Flag Capture]{
        \includegraphics[width=0.17\textwidth]{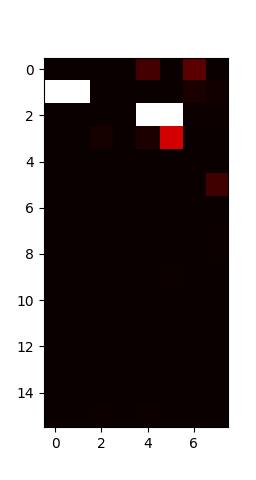}
        \label{fig:heatmap6}
    }
    \hfill 
    \subfloat[Entombed Competitive]{
        \includegraphics[width=0.17\textwidth]{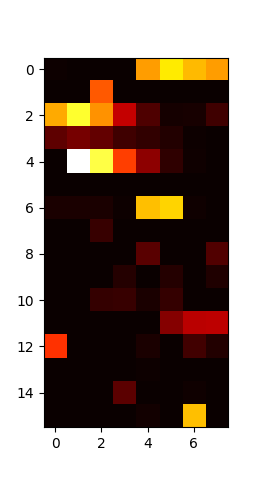}
        \label{fig:heatmap8}
    }
    \hfill 
    \subfloat[Entombed Cooperative]{
        \includegraphics[width=0.17\textwidth]{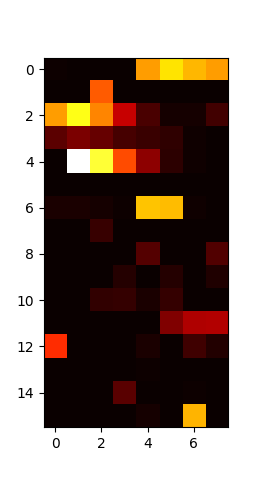}
        \label{fig:heatmap9}
    }
    \hfill 
    \subfloat[Surround]{
        \includegraphics[width=0.17\textwidth]{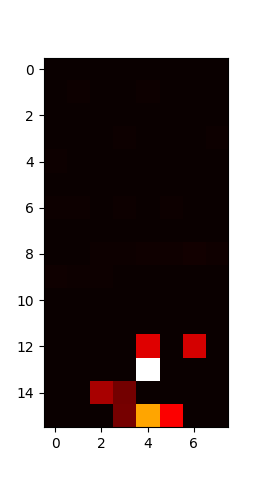}
        \label{fig:heatmap10}
    }
    
    \caption{Figure 2: Heatmap visualizations of RAM complexity for the ten games. Each pixel represents a RAM byte. A brighter color denotes higher temporal variations in that byte whereas a darker color denotes lower temporal variations.}
    \label{fig:heatmap}
\end{figure*}

\begin{table}[!ht]
  \begin{center}

    \begin{tabular}{c|c|c} 
    \hline 
      Game & From Scratch & Transferred \\

      \hline
      
      Double Dunk & 16.79 & 13.47 \\
      Mario Bros & 16.35 & 11.89 \\
      Space Invaders & 16.57 & 12.85 \\
      Tennis & 18.04 & 13.34 \\
      Boxing & 19.05 & 11.39 \\
      Pong & 16.62 & 12.44 \\
      Flag Capture & 16.08 & 12.25 \\
      Entombed Competitive & 15.14 & 12.28 \\
      Entombed Cooperative & 16.25 & 10.94 \\
      Surround & 16.36 & 11.07 \\
      
       \hline 
    \end{tabular}
  \end{center}
      \caption{Average run time (in hours) for the ten Atari games in the two-player versions.}
    \label{table:trainingtime}
\end{table}

\subsection{Average Run Time}

One advantage of transfer learning and layer freezing is the running time saved. Table \ref{table:trainingtime} shows the average time taken for each of the ten games in the two-player versions. In each case, the transferred version took less time than the version from scratch. This makes sense since layer freezing allows a network to be trained with less parameters. The results show that the transferred version took, on average, 27\% less time, a substantial saving.

\subsection{Average Total Rewards per Episode}
Figure 1 shows the average reward over five training sessions with different random seed values in transferred vs. from scratch settings, again both in the two-player versions. The graphs show a running average with a window of 10 and random sampling to present the trend of the training process more clearly. Table 2 presents the mean and standard deviation of average total rewards in five different points in time: At the start of training, 5000 episodes in, 10,000 episodes in, 15,000 episodes in, and lastly the final episode. These provide a sample of the training process for each agent in each game. 

In each graph in Figure 1, the blue curve indicates the transferred version while the orange curve indicates the version from scratch. It is interesting to observe that while in some games (in particular, Double Dunk, Space Invaders, and Tennis), the transferred version clearly outshines the version from scratch, this is not consistent across all games. While we do not observe any environments where the version from scratch is better than the transferred version, there are cases where they are closely matched. This addresses our first research question: we see that the transferred version is at least as good as the version from scratch in terms of rewards. However, it still begs the question: could the transferred agent performance be predicted with a simple predictor derived from the input (the Atari RAM)?

\begin{table*}[htb]
  \begin{center}

    \begin{tabular}{c|r|r} 
    \hline 
      Game & RAM & Normalized difference of means    \\
        & Complexity & of rewards of last 100 episodes  \\
      \hline
      Double Dunk & 454.15 & 0.646   \\
      Mario Bros & 394.42 & 0.255   \\
      Space Invaders & 328.22 & 0.468   \\
      Tennis & 285.66 & 0.518   \\
      Boxing & 217.84 & 0.588   \\
      Pong & 142.73 & -0.075   \\
      Flag Capture & 112.35 & 0.148   \\
      Entombed Competitive & 126.85 & 0.081   \\
      Entombed Cooperative & 127.54 & 0.121  \\
      Surround & 65.31 & 0.613  \\
    
     \hline 

    \end{tabular}
  \end{center}
      \caption{RAM complexity and normalized difference of means for each game.}
    \label{table:ram}
\end{table*}

\subsection{The Atari RAM Complexity}

There are different ways to provide a notion of ``complexity'' of the RAM usage. For our study, we focus on the temporal variation in the RAM data, as we feel that temporal information best captures a game in progress.
Variation indicates the change in each byte of the RAM, which can serve as an indicator of the RAM observation’s complexity.
More frequent changes in each byte suggest that specific bytes are more dynamic, reacting to different actions. 
Additionally, the emphasis is on temporal changes, as substantial changes in a byte value over time may be misleading when considered on a global scale. To mitigate this, RAM Complexity is calculated by analyzing specific time windows.

A dataset of 50,000 RAM instances is gathered by a random agent playing every two-player game. In each of these datasets, a difference is calculated between the actual data in each byte and average of the values that are neighbors of the data using a kernel of size 11 across the time dimension. Each difference value is squared. The average of these squared differences is considered the final value for a specific byte in the RAM values. We repeat this process for every byte in the RAM dataset and obtain values that show the temporal variation for each byte separately. We cap these values at a maximum of 3000 to reduce chances of outliers. As an Atari RAM has 128 bytes, this provides 128 separate values.

We present a method of visualizing the magnitude of RAM data variations in an Atari game environment. Using the 128 values gained from the previous step we can produce a heatmap for a game. We give this heatmap a dimension of 16 by 8 for better visualization. Figure 2 shows the resulting heatmaps. This visualization allows us to intuitively observe how differently each game utilized its RAM space - for some games, activities are concentrated in a few bytes, while in other games, activities are spread out in the RAM.

We present a measure for RAM complexity and analyze the Atari RAM of each game by this measure. From these 128 byte variation values, we can calculate one number that we define as our "RAM Complexity", by taking the average of the 128 values. The second column of Table \ref{table:ram} shows our defined RAM Complexity for each of the ten game environments. 

The last column of Table \ref{table:ram} shows a normalized difference of means based on the rewards of the last 100 episodes in each game. As the scales of the average total rewards vary by environment, a direct comparison across environments is not possible. Normalization involves rescaling the data to a common scale without distorting differences in the ranges of values, thus allowing for a comparison between the environments. To calculate the normalized difference of means, we first use a 90\% winsorization to remove extreme outliers in rewards. Then the rewards of the last 100 episodes in each environment are scaled to a [0, 1] range by performing a Min-Max normalization (using the minimum and maximum reward per each environment). Lastly, the difference between the mean of the scaled last 100 episodes rewards in transferred and non-transferred versions is calculated. 

We observe that generally there is a trend that as RAM complexity goes down, the advantage of the transferred agent goes down as well, with Surround being an obvious outlier. The Pearson Correlation Coefficient between the two sets of values is 0.44, indicating a weak positive relationship.  If the outlier Surround is removed, the Pearson Correlation Coefficient becomes a stronger 0.71. This suggests that a higher RAM complexity (based on our measure of temporal change) could be connected to the transferred agent performing better in this environment compared to agents trained from scratch, and that the benefits of transferred knowledge become less relevant in simpler environments. This addresses our second research question: in the absence of other information, we could use the RAM complexity as a weak predictor of the effectiveness of transfer. To our knowledge, we are the first to conduct this type of analysis.

\section{Conclusion, Limitations, and Future Work}
In this research, we investigate the potential of knowledge transfer from single-player to two-player Atari games as a solution to the prevalent issues of instability and computational inefficiency in training self-play two-player games. Through an empirical study conducted in the Atari environment, we demonstrate that transferred knowledge from a single-player setting leads to cumulative rewards that are at least as good as non-transferred settings, while reducing the total running time.

However, there are limitations to our results. We have chosen ten environments with one and two-player versions, and these games all have the player avatar(s) shown on the screen so that our agent indication method can annotate an avatar's location. We have not tested our work on games that do not have an on-screen player avatar (such as Video Checkers). Moreover, most Atari games have symmetric graphics largely due to the technical limitations of the system's hardware. While we use the Atari RAM as the input, the symmetry in the level design of some games could have effects that are not captured by the current results.

Hardware constraints may affect the choice of the number of training steps, thus potentially preventing the agents from achieving optimal performance. A comprehensive investigation into the advantages of knowledge transfer may be facilitated by extending the training duration until the agents reach a predetermined performance level.

For the purpose of this study, DQN was chosen due to its prevalence in Atari games. However, using different algorithms can produce varying transfer results. Future research could explore the differences between using different type of algorithms, i.e. off-policy and on-policy algorithms for knowledge transfer.

Atari games serve as a prime example of how an agent's performance can be enhanced by transferring knowledge from single-player versions. The implications of knowledge transfer warrant further investigation in diverse environments, particularly those exhibiting varied characteristics. This includes, but is not limited to, environments that are wholly collaborative rather than competitive, or those where the observations in the single-player and multiplayer versions manifest significant differences.

Our research also provides a method of analysis of the Atari RAM, including a visualization of its complexity by using a temporal change measure. While we show that there is a weak correlation between RAM complexity and the performance of the transferred agent, there is still much to be done. Further research should be conducted to validate the accuracy and effectiveness of using RAM complexity as a predictor for agent performance.

\section{Acknowledgements}
This research was supported by the Natural Sciences and Engineering Research Council of Canada (NSERC) Discovery Grant. We thank members of the Serious Games Research Group and the anonymous reviewers for their feedback.

\bibliography{aaai24}

\end{document}